\documentclass[letterpaper, 10 pt, conference]{ieeeconf} 

\IEEEoverridecommandlockouts
\usepackage{amsmath,amssymb,amsfonts}
\usepackage{algorithmic}
\usepackage{graphicx}
\usepackage{textcomp}
\usepackage{xcolor}
\usepackage{subcaption}
\usepackage{wrapfig}
\usepackage{hyperref}
\usepackage{multirow}

\definecolor{red}{rgb}{1.00,0.00,0.00}
\definecolor{blue}{rgb}{0.00,0.00,1.00}
\definecolor{green}{rgb}{0.30, 0.50,0.00}

\def\BibTeX{{\rm B\kern-.05em{\sc i\kern-.025em b}\kern-.08em
    T\kern-.1667em\lower.7ex\hbox{E}\kern-.125emX}}
    
\begin{document}

% \title{Reinforcement Learning for Robotic Object Throwing with Obstacle Avoidance}
% \title{Mastering Object Throwing in Multi-Obstacle Environments through Reinforcement Learning}
\title{Learning to Throw Objects Safely in Multi-Obstacle Environments}
\author{Mohammadreza Kasaei$^{2}$, Klemen Voncina$^{1}$, and Hamidreza Kasaei$^{1}$
\thanks{$^{1}$ Klemen Voncina and Hamidreza Kasaei are with the Department of Artificial Intelligence, University of Groningen, The Netherlands. Emails: k.voncina@student.rug.nl, hamidreza.kasaei@rug.nl}
\thanks{$^{2}$ Mohammadreza Kasaei is with the School of Informatics, University of Edinburgh, UK. Email: m.kasaei@ed.ac.uk}%
}

\maketitle

\begin{abstract}
Robotic throwing enables fast and efficient object placement beyond the robot’s immediate workspace, but reliable throwing in cluttered environments remains underexplored. Existing approaches, such as TossingBot, learn throwing strategies from visual input but assume obstacle-free settings. In this paper, we address the problem of throwing objects into a target basket while avoiding obstacles placed randomly in the scene. We introduce a potential field state representation that compactly encodes both basket attraction and obstacle repulsion on a fixed-size grid, enabling reinforcement learning (RL) policies to generalize across arbitrary numbers and configurations of obstacles. The policy is initialized from kinesthetic demonstrations and optimized in simulation using three state-of-the-art RL algorithms (SAC, DDPG, TD3). Among these, SAC achieves the most consistent performance across scenarios. We compare the potential field representation against explicit state encodings and demonstrate that it achieves higher success rates and better scalability to unseen obstacle configurations. Real-robot experiments with unseen throwable objects confirm robust sim-to-real transfer, achieving up to $90\%$ success in cluttered scenes. These results demonstrate that PFR provides a practical and robust representation for safe and efficient robotic throwing in unstructured environments. A video showcasing our experiments is available at: \href{https://youtu.be/ZZnJf8ua2dE}{\texttt{\small https://youtu.be/ZZnJf8ua2dE}}

\end{abstract}

\section{Introduction}
In automated distribution and sorting logistics systems, speed and efficiency play a critical role \cite{zermane2024planning}. 
Developing robots with the ability to swiftly pick up and throw objects can significantly enhance throughput and operational efficiency.
Throwing, in particular, enables fast and precise object placement, reduces the need for complex robotic traversal, and extends the robot's effective workspace \cite{raptopoulos2020robotic}. 
By allowing robots to toss objects instead of carrying them, throughput can be improved in warehouses, automated fulfillment centers, and waste-sorting systems.

Recent works have demonstrated the feasibility of robotic throwing. 
For example, TossingBot~\cite{zeng2019tossingbot} learns grasping and throwing policies from visual input, but assumes obstacle-free environments. 
Other approaches~\cite{kasaei2023throwing,rl_adapt} rely on hand-crafted or analytic motion kernels, limiting adaptability in cluttered settings. 
However, practical environments rarely allow direct throwing trajectories: obstacles such as walls, bins, or other items must be considered to ensure successful placement.

\begin{figure}[!t]
  \centering
  \includegraphics[width=\linewidth]{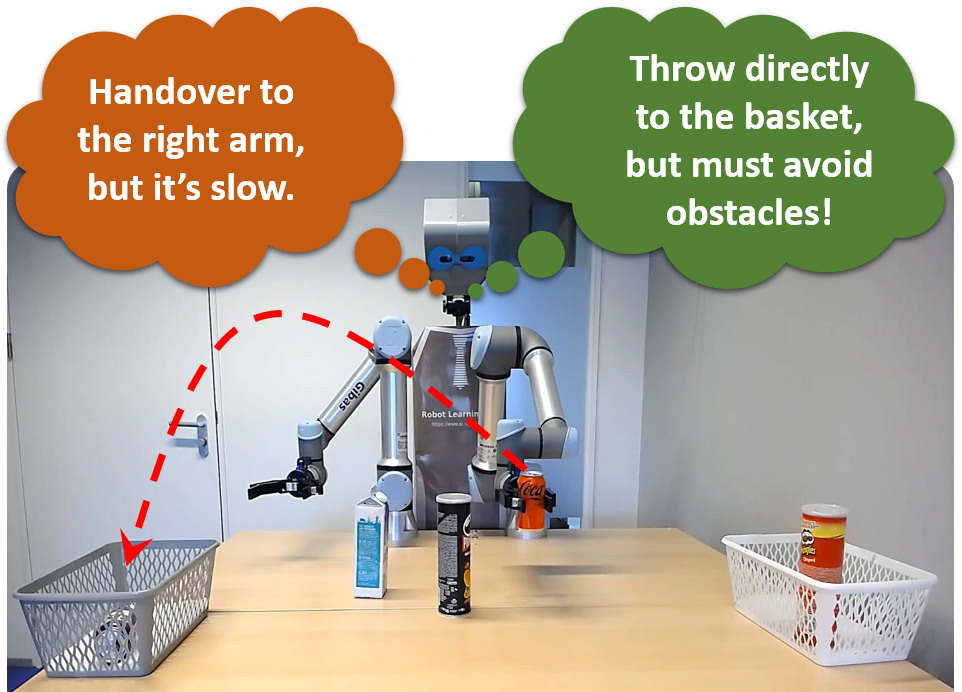}
\caption{Overview of our throwing task. The robot must decide between a slow object-handover or a faster throwing action. Since obstacles may block the trajectory, we treat the problem as a \textbf{Safe Reinforcement Learning} challenge, where the goal is to achieve reliable throws while avoiding collisions.}
  \label{fig:task_overview}
\end{figure}

An illustrative example is shown in Fig.~\ref{fig:task_overview}. 
Suppose the robot is asked to place an object into the gray basket. 
Since the basket lies outside the maximum kinematic reach of the left arm, a simple placement action is not feasible. 
One option would be to perform a hand-over to the right arm and then place the object into the basket, but this is a relatively slow process and reduces system throughput. 
Alternatively, the robot can throw the object directly into the basket. 
However, this introduces additional challenges: the robot must decide whether a safe throwing trajectory exists in the presence of obstacles, and if so, how to execute it reliably. 
This motivates formulating the problem as a safe reinforcement learning task, where the robot learns to maximize throwing success while avoiding collisions.

In this paper, we address the problem of throwing objects into a target basket in the presence of multiple, randomly placed obstacles. We introduce a {potential field state representation} that compactly encodes both basket attraction and obstacle repulsion on a fixed grid. Unlike explicit state encodings that scale poorly with the number of obstacles, the potential field provides a fixed-dimensional and physically meaningful input to the learning algorithm, enabling policies to generalize across arbitrary numbers and configurations of obstacles.

To avoid unsafe random exploration in cluttered scenes, we initialize the throwing motion using kinesthetic teaching~\cite{akgun2012trajectories}. 
These demonstrations provide a safe kernel that bootstraps reinforcement learning and ensures stable early training. The kernel is then modulated by policies trained with reinforcement learning. We compare three state-of-the-art RL algorithms (SAC, DDPG, TD3) through this project and analyze their performance on the throwing task. The key contributions of our work are:
\begin{itemize}
    \item We formulate robotic throwing with obstacles as a safe reinforcement learning problem, where the robot must learn to throw while avoiding collisions (Fig.~\ref{fig:task_overview}).
    We leverage kinesthetic teaching to initialize the throwing kernel safely, reducing unsafe exploration during early RL training.
    \item We introduce a compact potential field state representation that encodes basket and obstacle information in a fixed grid, enabling scalable throwing policies that generalize across arbitrary numbers of obstacles.
    \item We conduct extensive experiments in both simulation and on a real robot, showing that policies trained with potential fields achieve high success rates and transfer effectively to unseen objects and obstacle configurations.
    % \item We benchmark three popular RL algorithms (SAC, DDPG, TD3) for this task and show that SAC consistently achieves the best overall performance.
\end{itemize}

\section{Related work}

% The problem of robotic object throwing has been widely explored using both analytical methods and learning-based approaches~\cite{bombile2022dual,forrai2023event,bombile2023bimanual,liu2022solution,monastirsky2022learning}. In this section, we review prior works, highlighting their limitations and positioning our contribution within this landscape.

Robotic throwing has been studied through a variety of approaches, ranging from analytic motion planning and physics-based models to modern reinforcement learning methods~\cite{bombile2022dual,forrai2023event,bombile2023bimanual,liu2022solution,monastirsky2022learning}. 
Analytical approaches typically rely on explicit models of object dynamics and release kinematics, enabling precise but brittle solutions. 
Learning-based methods, on the other hand, allow robots to adapt to unknown dynamics and unmodeled effects, but most prior works focus on obstacle-free scenarios or encode obstacles explicitly in the state, which does not scale to complex environments. 

In this section, we first review analytical methods for throwing and highlight their limitations in unstructured settings. We then discuss learning-based methods, emphasizing recent works that combine reinforcement learning with kernel modulation. Finally, we position our contribution within this landscape.

\subsection{Analytical methods}

These approaches aim to explicitly model object dynamics, throwing kinematics, and release timing to generate precise throwing trajectories. Early works focused on deterministic motion planning and rigid-body dynamics models for controlled throwing.

For instance, sampling-based motion planning methods have been successfully applied to throwing tasks. Zhang et al.~\cite{zhang2012sampling} introduced dynamic intermediate state objectives to plan throwing motions effectively. Similarly, LaValle and Kuffner\cite{lavalle2001randomized} and Kunz and Stilman~\cite{kunz2014probabilistically} proposed kinodynamic planning techniques that account for velocity and acceleration constraints, making them suitable for throwing tasks.

More recently, adaptive and impact-driven planning methods have emerged. Liu et al.~\cite{liu2022solution} developed a model for adaptive mobile manipulator throwing, while Zermane et al.~\cite{zermane2024planning} proposed impact-driven logistic planning to optimize object release timing. Heavily constraining the problem, such as proposed by Myashita et al.~\cite{one_joint}, may also help to restrict the solution space sufficiently to approach the problem analytically. 

Although these analytical methods can achieve high accuracy, they struggle in dynamic and unstructured environments, where real-world uncertainties (e.g., object properties, obstacle variations, and sensor noise) are difficult to model explicitly ~\cite{lombai2009throwing, dyn_man,gai2013motion,kim2016executing}. For example~\cite{dyn_man} describes a throwing/catching robot that uses padded manipulators to avoid having to deal with the complex dynamics of a ball bouncing off a hard surface when being caught. In~\cite{one_joint}, the robotic manipulator consists of a single joint to constrain the degrees of motion that need to be accounted for. Even~\cite{zeng2019tossingbot} constrain the parameter kernel by fixing the release height of the robot, therefore only having to deal with parameters for speed, angle, and release timing. Schwarke et al. introduced a curiosity-driven learning approach for joint locomotion and manipulation tasks, where in an obstacle-free environment, their robot performs a joint location and manipulation task involving throwing a package from a table to a bin~\cite{schwarke2023curiosity}. Our work differs from these approaches by leveraging deep reinforcement learning (RL), which can adapt to such uncertainties without requiring handcrafted models.

% \subsection{Learning methods}
% To address the limitations of analytical methods, researchers have explored data-driven and reinforcement learning-based throwing.
% Kober et al.~\cite{rl_adapt} propose a method that combines reinforcement learning and physical modeling. A significant breakthrough in RL-based throwing came with TossingBot~\cite{zeng2019tossingbot}, which integrates residual physics modeling into an end-to-end learning framework. This allows the robot to learn object dynamics implicitly, reducing dependence on predefined models. However, TossingBot primarily focuses on obstacle-free scenarios, limiting its applicability in multi-obstacle environments. More recent RL-based works have tackled obstacle-avoiding throwing. Kasaei et al., ~\cite{kasaei2023throwing} introduced an RL framework for throwing into a moving basket while avoiding an obstacle, but their method does not consider user preferences and multi-obstacle scenarios. Our approach extends these works by incorporating kinesthetic teaching to allow the user to influence the throwing policy and evaluate multi-object fine-tuning, showing how RL models adapt (or fail to adapt) to unseen object dynamics.

\subsection{Learning methods}
To address the limitations of analytical methods, researchers have explored data-driven and reinforcement learning-based throwing.
Kober et al.~\cite{rl_adapt} propose a method that combines reinforcement learning and physical modeling. 
A significant breakthrough in RL-based throwing came with TossingBot~\cite{zeng2019tossingbot}, which integrates residual physics modeling into an end-to-end learning framework. 
This allows the robot to learn object dynamics implicitly, reducing dependence on predefined models. 
However, TossingBot primarily focuses on obstacle-free scenarios, limiting its applicability in cluttered environments.

More recent RL-based works have considered throwing with obstacles. 
Kasaei et al.~\cite{kasaei2023throwing} introduced an RL framework for throwing into a moving basket while avoiding a single obstacle. 
In their approach, the pose of the obstacle is explicitly included in the state vector. 
While effective for a fixed number of obstacles, this design is not scalable: each additional obstacle increases the state dimension and requires training a new policy for that specific configuration. 
In contrast, our method employs a compact potential field representation that encodes both basket attraction and obstacle repulsion in a fixed-size grid, allowing a single policy to generalize across arbitrary numbers and configurations of obstacles.

Our approach further differs by leveraging kinesthetic teaching to safely initialize the throwing kernel. 
Rather than requiring the policy to explore from scratch, demonstrations provide a safe starting point for reinforcement learning in cluttered scenes. 
We systematically evaluate the scalability of this approach across varying numbers of obstacles and show transfer to unseen objects, highlighting the robustness of our framework.

%%%%%%%%%%%%%%%%%%%%%%%%%%%%%%%%%%%%%%%%%%%%%%%%%%%%%%%%%%%%%%%%%%%%%%%%%
\section{Methods}\label{sec:methods}

\subsection{Preliminaries and Problem Formulation}
We formulate robotic throwing as a Markov Decision Process (MDP) with continuous state and action spaces. 
An MDP is defined by the tuple $(s_t, a_t, p(s_{t+1}|s_t,a_t), r(s_t,a_t))$, where $s_t$ is the state at time $t$, $a_t$ the action taken, $p(s_{t+1}|s_t,a_t)$ the transition dynamics, and $r(s_t,a_t)$ the reward. 
The objective of reinforcement learning is to find a policy $\pi(a|s)$ that maximizes the expected discounted return $R_t=\mathbb{E}[\sum_{i=t}^\infty \gamma^{i-t} r_{i+1}]$ with discount factor $\gamma \in [0,1]$. Since our action space is continuous and directly parameterizes the throwing kernel, we employ off-policy algorithms designed for continuous control: Deep Deterministic Policy Gradient (DDPG)~\cite{lillicrap2015continuous}, Twin Delayed DDPG (TD3)~\cite{fujimoto2018addressing}, and Soft Actor-Critic (SAC)~\cite{haarnoja2018soft}. 
These algorithms reuse samples from a replay buffer to improve sample efficiency, making them well-suited for robotic domains where interactions are costly. 
% In practice, SAC generally provides the most stable performance due to its entropy-maximization objective, while DDPG and TD3 offer useful baselines for comparison.

We study the task of throwing an object into a target basket in the presence of randomly placed obstacles. The policy receives an observation of the scene and must output throwing parameters that modulate a kernel initialized by kinesthetic teaching. Each episode consists of one throwing attempt: the object is placed in front of the robot, a basket is placed at a random reachable location, and between 0--5 obstacles are placed in the workspace (Fig.~\ref{fig:env_all_example}). 
The policy is rewarded if the throw lands successfully in the basket without colliding with obstacles.

\subsection{Kinesthetic Teaching for Safe Initialization}
\begin{wrapfigure}{r}{0.45\linewidth}
\vspace{-3mm}
    \includegraphics[width=\linewidth]{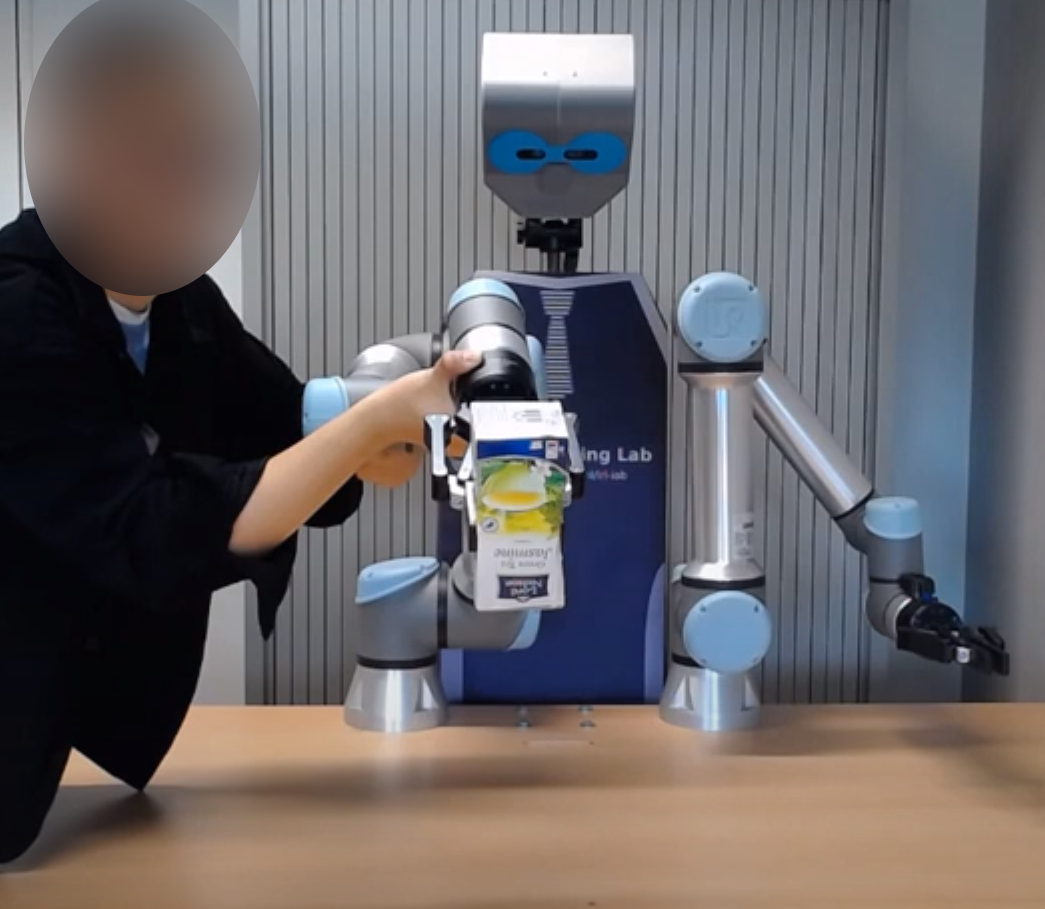}
    \caption{Kinesthetic teaching of the throwing kernel. 
    A human demonstrator physically guides the robot arm through a throwing motion. 
    This provides a safe initial kernel that bootstraps RL training and avoids unsafe random exploration.}
    \label{fig:kinesthetic_teaching}
    \vspace{-2mm}
\end{wrapfigure}
To avoid unsafe random exploration at the start of RL training, we use kinesthetic teaching~\cite{akgun2012trajectories}. 
A human demonstrator physically guides the robot arm through a throwing motion, which defines an initial \textit{throwing kernel} (see Fig.~\ref{fig:kinesthetic_teaching}). 
This kernel encodes joint trajectories that are safe and physically feasible, serving as a structured prior.  The RL policy does not imitate the demonstration directly, but instead learns to modulate kernel parameters---initial and final shoulder joint values, release time, and throw duration---to adapt to different objects and obstacle configurations. Thus, kinesthetic teaching is not used for personalization, but as a mechanism for safe exploration.

\begin{figure}[!t]
% \begin{wrapfigure}{r}{0.55\linewidth}
% \vspace{-3mm}
    \includegraphics[width=\linewidth, trim={0cm, 0cm, 0cm, 0cm}, clip=true]{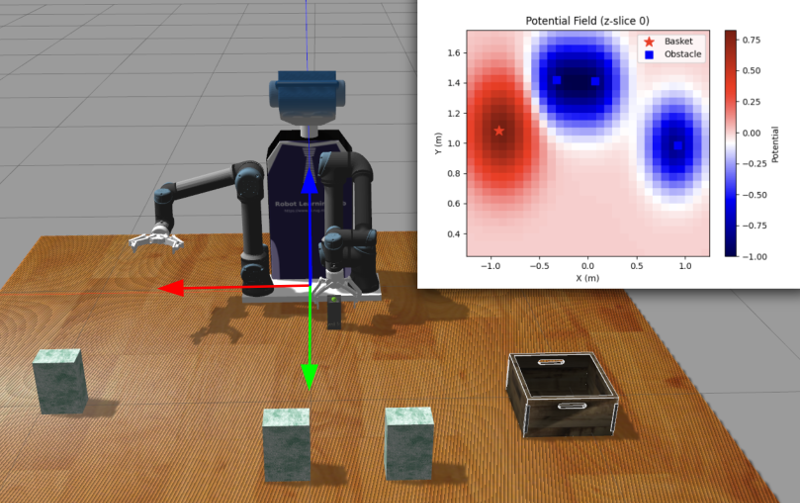}
\caption{Example of the Potential Field Representation (PFR). 
The robot’s workspace is discretized into a fixed grid, where the basket generates an attractive potential (red) and obstacles generate repulsive potentials (blue). 
This yields a structured, fixed-size encoding that generalizes across different numbers of obstacles.}
    \label{fig:PFR}
\vspace{-3mm}
% \end{wrapfigure}
\end{figure}
\subsection{State Representation}
The state representation combines both robot-centric variables and task-level information. 
At its core, each state includes a \textit{proprioceptive block} consisting of the initial and final values of the shoulder joint $(j_i,j_f)$, and a \textit{timing block} capturing the gripper release time $t_r$ and the overall motion duration $\theta$. 
Together, these four variables describe the kernel parameters that directly modulate the throwing action. 
In addition, the state contains the position of the throwable object $(x_o,y_o,z_o)$ and the basket $(x_b,y_b,z_b)$ expressed in the robot base frame, as well as their relative distances $(d, d_x, d_y)$. Including explicit distances accelerates learning by allowing the policy to reason about spatial relationships without inferring them solely from raw coordinates.

\begin{figure*}[!t]
    \centering
    \includegraphics[width=\textwidth]{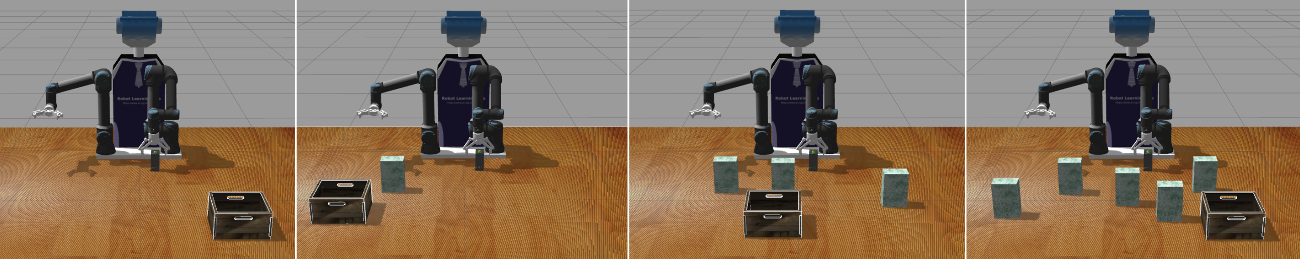} 
    \caption{Examples of environment configurations: no obstacles (left), 1 obstacle, 3 obstacles, and 5 obstacles (right). 
    The robot must learn which obstacles block the throwing trajectory and which do not.}
    \label{fig:env_all_example}
\end{figure*}

The main difference between the variants of our method lies in how obstacles are represented. 
In the \textbf{Explicit Pose Representation (EPR)}, each obstacle is encoded by appending its center position and its relative distance to the robot frame. This produces a compact, low-dimensional observation vector that allows fast convergence, but its dimensionality grows linearly with the number of obstacles, making it suitable for scenarios with known obstacles. Therefore, we construct an \textit{observation space} for each of our test conditions. For example, the state of an environment containing the throwable, the goal object, and no obstacles would look like this $s = (proprio,timing,pos_o,pos_g,dst_g) \in \mathbb{R}^{13}$. To this basic state, we can then add the representation of an obstacle as well. Thus, this environment with one obstacle would record a state in the following format; \mbox{$s = (proprio,timing,pos_o,pos_g,dst_g,pos_{obs},dst_{obs}) \in \mathbb{R}^{19}$} any further obstacles each add 6 more observable values to the state. Of course, these variables are bounded by certain constraints. All the positional variables will appear within the pre-determined range of the robot's workspace.

In contrast, the \textbf{Potential Field Representation (PFR)} encodes all obstacles and the basket jointly in a fixed-size grid covering the robot’s workspace (see Fig.~\ref{fig:PFR}). The workspace is discretized into an $M \times M$ grid (with $M=15$). 
Each cell stores a continuous potential value: attraction from the basket (positive) and repulsion from obstacles (negative), decaying smoothly with distance. This produces a structured representation whose dimensionality is independent of the number of obstacles, allowing a single policy to generalize across arbitrary obstacle counts and configurations. We then introduce a convolutional neural network (CNN) encoder for the PFR. 
Instead of flattening, the potential field is treated as a single-channel image of size $(1, M, M)$, which is processed by a CNN backbone to extract spatial features. These features are then concatenated with proprioceptive variables (joint values, release time, and throw duration) before being passed to the policy network. This design preserves the geometric structure of the environment, enabling the policy to more effectively exploit the smooth gradients of the potential field and the shape of obstacles. 
While PFR requires more training interactions due to its higher dimensionality, it provides scalability and a physically meaningful encoding that supports policies trained once to be deployed across arbitrary numbers of obstacles.

 % We first flattened the potential field into a 1D vector ($M^2$ dimension) and concatenated it with proprioceptive features before passing it to the policy and value networks. 

\subsection{Action Space}
The action space directly parameterizes the throwing kernel that the robot executes. Each action is a four-dimensional continuous vector, $a_t = (j_i, j_f, t_r, \theta)$, where $j_i$ and $j_f$ denote the initial and final values of the shoulder joint, $t_r$ is the release time of the gripper, and $\theta$ is the duration of the throwing motion. 
Given these parameters, the controller generates a trajectory in which the shoulder joint moves from $j_i$ to $j_f$, the second joint extends with a fixed amplitude provided by the kinesthetic demonstration, and the gripper opens at $t_r$ within the total motion duration $\theta$.  This formulation allows RL to modulate a safe kernel initialized from demonstration rather than learning raw joint trajectories from scratch, thereby reducing the risk of unsafe exploration and focusing the search on physically meaningful throwing motions.

\subsection{Environment Setup}
A simulated environment is implemented using Gazebo and ROS~\cite{quigley2009ros,macenski2022robot}, which mirrors the real robot with two UR5e arms and an Xtion vision system mounted on the head of the robot. Each training episode consists of a single step, therefore, two observations (before and after) and one action. In each experiment, an object is placed in front of the robot to be thrown. The goal basket is placed randomly in the robot's vicinity. If there are obstacles in the environment, they are placed randomly in the scene, ensuring that they do not collide with each other, the goal basket, or a throwable object. {The random placement of obstacles, even when some may appear distant from the basket or outside the robot's immediate workspace, serves a purpose in our study (refer to Fig.~\ref{fig:env_all_example}, the last two images on the right). It allows the robot to discern and learn which objects should be considered obstacles preventing the throwing trajectory toward the target basket and which objects are not obstructive to reaching the target position. By exposing the system to various spatial configurations, including those where certain obstacles seem less relevant, we enable the robot to develop a nuanced understanding of the relevant obstacles and refine its decision-making process during the learning phase.
} 

\subsection{Reward}\label{sec:reward}
The reward for the action is calculated based on the state as recorded at the end of the training episode. Firstly, a few conditions are checked. If an obstacle has been moved (caused by collision) the reward is set to 0, regardless of whether it reaches the goal, if the object is inside the goal box (checked by distance and axis-aligned bounding box collision calculation), the reward is 1, otherwise, the reward is calculated using the following function:
\begin{equation}\label{eq:reward}
    r = \rho - \sigma
\end{equation} 

\noindent where the reward term $\rho$ is described by
\begin{equation}
    \rho = 0.9 e^{-100 d_g^2} + 0.1 e^{-2d_g^2}
\end{equation}

\noindent where the choice of exponents $e^{-100}$ and $e^{-2}$ ensures rapid decay for large errors and allows gradual improvement, respectively. The 
$e^{-100}$ term decays extremely fast, meaning that a throw far from the goal gets almost no reward, strongly discouraging bad throws, while the $e^{-2}$ term decays more slowly, meaning even suboptimal throws receive some reward, helping the policy refine its performance incrementally. This approach prevents the reward from becoming sparse. The penalty term $\sigma$ is given by
\begin{equation}
    \sigma = (1 - e^{-{10 d_r^2}})
\end{equation}

\noindent $d_g$ is the same distance that appears in the observation between the object and obstacle at the final position of the training episode and $d_r$ is the relative distance of the object to the goal measured as the difference between $d_g$ at the start of the episode and $d_g$ at the end of the episode. Additionally, should this term be positive (the object is closer than it was at the start), $d_r = 0$. 

This ensures that we only penalize the throw if the object ended up further away after the throwing action, meaning the robot threw it in a random direction.

\section{Experiments and Results}

We validate our approach through extensive experiments in both simulation and on a real robotic platform. 
Our evaluation has two main objectives: (i) to compare two alternative obstacle encodings, the Explicit Pose Representation (EPR) and the Potential Field Representation (PFR), in terms of learning efficiency and scalability, and (ii) to assess the generalization of learned throwing policies to unseen objects and their transferability from simulation to the real robot. 
Performance is measured in terms of the \textit{throwing success rate}, defined as the ratio of successful throws that land inside the basket to the total number of attempts.

\subsection{Training Setup}
Policies are trained in simulation using the milk box object as the primary throwable and then evaluated on both the milk box and three previously unseen objects (banana, coke can, and sneaker). 
For EPR, policies are trained for $100$k timesteps, while PFR requires $250$k timesteps due to its higher-dimensional observation space. 
We consider four obstacle scenarios with $0, 1, 3,$ and $5$ obstacles randomly placed in the workspace (Fig.~\ref{fig:env_all_example}). 
Training in simulation allows safe and efficient exploration, after which the best-performing policies are transferred zero-shot to the real robot. 

Our implementation is based on Stable-Baselines3~\cite{stable-baselines3}, and Table~\ref{tab:hyperparam} summarizes the hyperparameters used across SAC, TD3, and DDPG. 
The perception pipeline for real-robot tests (Fig.~\ref{fig:real_perception}) builds on prior work in object detection and grasp synthesis~\cite{kasaei2023mvgrasp,kasaei2021simultaneous,kasaei2019interactive,kasaei2018towards}.

\subsection{Ablation on Potential Field Grid Resolution}
We further investigated the effect of grid resolution in the Potential Field Representation (PFR). 
In this experiment, we compared three grid sizes: $10 \times 10$, $15 \times 15$, and $30 \times 30$.

\begin{figure}[!b]
    \vspace{-5mm}
    \centering
    \includegraphics[width=\linewidth]{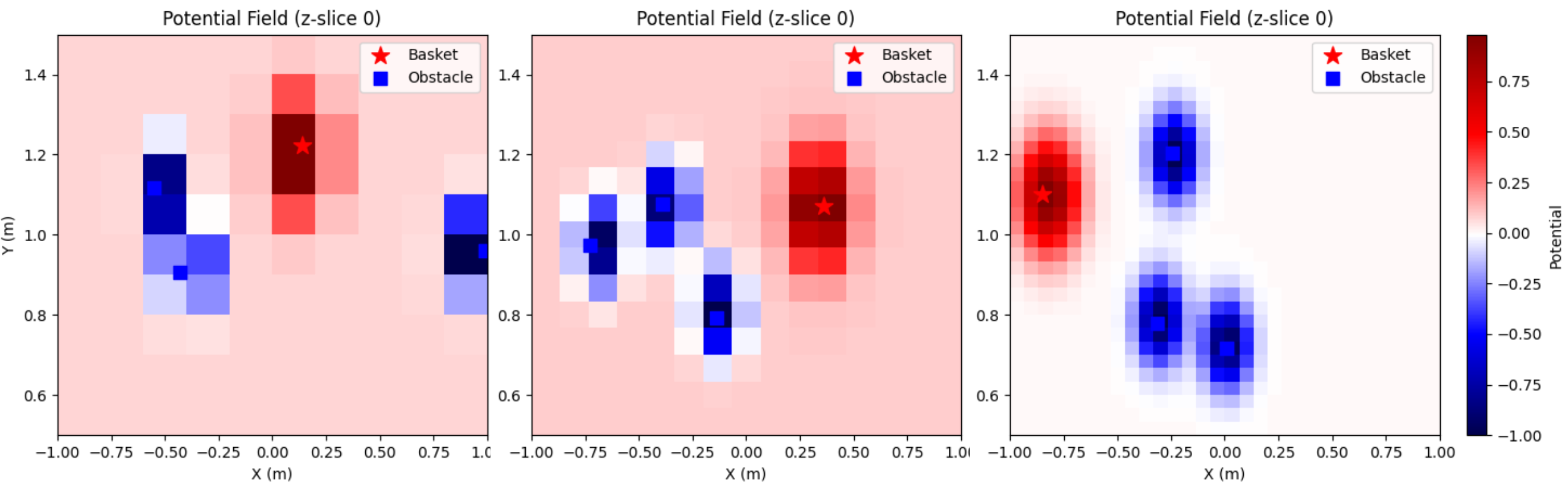}
\caption{Effect of grid resolution on the Potential Field Representation (PFR). 
(\textit{left}) A $10 \times 10$ grid provides a coarse and blocky representation, 
where small obstacles occupy only a few cells. 
(\textit{middle}) A $15 \times 15$ grid offers a clearer encoding of basket attraction and obstacle repulsion. 
(\textit{right}) A $30 \times 30$ grid produces smooth gradients and sharper obstacle boundaries, 
capturing fine details at the expense of increased training cost.}
    \label{fig:grid_ablation}
\end{figure}

Figure~\ref{fig:grid_ablation} illustrates representative potential fields at two different resolutions. 
With a coarse $10 \times 10$ grid, the field is blocky, and small obstacles are represented by only a few cells, 
making their influence less precise. Increasing the resolution to $15 \times 15$ produces smoother gradients, 
while a $30 \times 30$ grid yields even finer details, capturing obstacle boundaries and basket attraction regions more clearly.  
\begin{table}[!t]
    \centering    
    \caption{List of hyper-parameters used in this study. For the CNN-based PFR, the potential field grid 
    is treated as a single-channel image and passed through a two-layer convolutional encoder before concatenation 
    with proprioceptive features.}
    \resizebox{0.8\linewidth}{!}{
    \begin{tabular}{|c|c|}
    \hline
    \textbf{Parameter} & \textbf{Value} \\
    \hline\hline
\#hidden layers (actor/critic MLPs) &  $2$ \\
\#hidden units per layer & $256$\\
\#samples per minibatch & $256$\\
optimizer & Adam\\
learning rate  & $3 \times 10^{-4}$\\
batch size (SAC) & 256 \\
batch size (TD3) & 100 \\
batch size (DDPG) & 100 \\
\#epochs & 50K \\
discount ($\gamma$)  & $0.99$\\
replay buffer size &  $50K$\\
nonlinearity & ReLU\\
target update rate ($\tau$) & 0.005\\
target update interval & 1 \\
gradient steps & 1 \\
\hline\hline
\multicolumn{2}{|c|}{\textbf{CNN Encoder (for PFR)}} \\ \hline
input channels & $1$ (potential field) \\
Conv layer 1 & $16$ filters, $3\times3$ kernel, stride $1$, ReLU \\
Conv layer 2 & $32$ filters, $3\times3$ kernel, stride $1$, ReLU \\
Flatten & fully connected + concat with proprioception \\
final linear layer & 256 units, ReLU \\
\hline
    \end{tabular}}
    \label{tab:hyperparam}
    \vspace{-5mm}
\end{table}

Quantitatively, we observe that increasing the grid size improves generalization to unseen objects and cluttered environments. 
For example, with SAC in the three-obstacle scenario, success rates for the milk box improve from $0.89$ with a $10 \times 10$ grid 
to $0.96$ with a $15 \times 15$ grid, and $0.97$ with a $20 \times 20$ grid. 
However, the computational cost also increases: the $10 \times 10$ grid converges after approximately $200$k steps, while the $20 \times 20$ grid requires closer to $300$k steps for stable performance.  These results reveal a trade-off between training efficiency and spatial fidelity. Coarser grids train faster but lose fine-grained obstacle information, whereas denser grids yield smoother and more discriminative potential fields at the expense of longer training. 
In practice, we find the $15 \times 15$ grid offers the best balance between efficiency and accuracy, and we adopt this setting in subsequent experiments.

\begin{figure*}[!t]
    \centering
     \includegraphics[width=\textwidth, trim={0.35cm, 0.5cm, 0.35cm, 0cm}, clip=true]{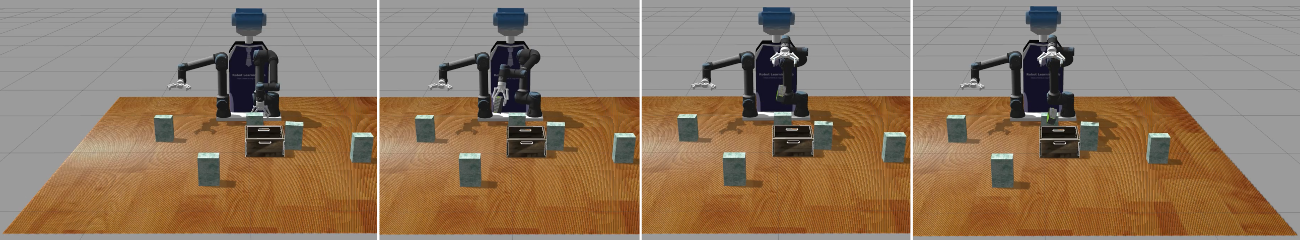} 
    \caption{Sequence of snapshots showing the policy successfully delivers the object into the target basket while navigating around multiple obstacles, demonstrating its ability to generate safe and effective throws in cluttered environments.}
    \label{fig:env_all_exampl}
    \vspace{-3mm}
\end{figure*}

\subsection{Simulation Results}
Table~\ref{tab:init_results} reports the performance of policies trained on the milk box when evaluated on both the trained object and three unseen objects across different obstacle scenarios. Among the algorithms, SAC consistently outperforms TD3 and DDPG, maintaining success rates above $90\%$ 
with up to three obstacles and showing greater robustness as clutter increases. 
In contrast, TD3 accuracy drops sharply in heavily cluttered scenes, while DDPG exhibits inconsistent generalization across objects. Interestingly, despite its elongated shape, the banana transfers relatively well under SAC, whereas the soda can and sneaker prove more challenging, reflecting the difficulty of generalizing to objects with different mass distributions and centers of gravity. 
\begin{table}[!t]
\caption{Throwing accuracy for the proposed model. 
Bold indicates the trained object; others are unseen test objects.}
\resizebox{\linewidth}{!}{
\begin{tabular}{|l|c|cc|cc|cc|cc|}
\hline
\multirow{2}{*}{Alg.} & \multirow{2}{*}{$\#$Obst} 
& \multicolumn{2}{c|}{\textbf{Milk}} 
& \multicolumn{2}{c|}{Banana} 
& \multicolumn{2}{c|}{Coke Can} 
& \multicolumn{2}{c|}{Sneaker} \\ \cline{3-10}
 & & EPR & PFR & EPR & PFR & EPR & PFR & EPR & PFR \\ \hline \hline
SAC & 0 & 0.95 & 0.97 & 0.91 & 0.94 & 0.85 & 0.93 & 0.94 & 0.94 \\
SAC & 1 & 0.95 & 0.97 & 0.92 & 0.94 & 0.81 & 0.91 & 0.91 & 0.93 \\
SAC & 3 & 0.93 & 0.96 & 0.83 & 0.90 & 0.76 & 0.90 & 0.90 & 0.93 \\
SAC & 5 & 0.86 & 0.92 & 0.75 & 0.83 & 0.73 & 0.82 & 0.82 & 0.87 \\ \hline \hline
TD3 & 0 & 0.92 & 0.93 & 0.72 & 0.89 & 0.71 & 0.89 & 0.84 & 0.89 \\
TD3 & 1 & 0.87 & 0.91 & 0.74 & 0.86 & 0.43 & 0.83 & 0.78 & 0.81 \\
TD3 & 3 & 0.79 & 0.84 & 0.66 & 0.86 & 0.35 & 0.75 & 0.71 & 0.81 \\
TD3 & 5 & 0.70 & 0.79 & 0.48 & 0.81 & 0.40 & 0.75 & 0.57 & 0.76 \\ \hline \hline
DDPG & 0 & 0.88 & 0.91 & 0.50 & 0.65 & 0.51 & 0.72 & 0.79 & 0.84 \\
DDPG & 1 & 0.92 & 0.92 & 0.67 & 0.67 & 0.58 & 0.67 & 0.83 & 0.84 \\
DDPG & 3 & 0.92 & 0.90 & 0.70 & 0.68 & 0.70 & 0.67 & 0.88 & 0.80 \\
DDPG & 5 & 0.83 & 0.87 & 0.67 & 0.68 & 0.69 & 0.68 & 0.74 & 0.81 \\ \hline
\end{tabular}}
\label{tab:init_results}
\vspace{-5mm}
\end{table}

A key result is that the PFR often surpasses the EPR. For example, with SAC and three obstacles, PFR reaches $0.90$ accuracy on the Coke Can compared to $0.76$ with EPR. Similarly, under TD3, PFR improves performance on the Banana from $0.66$ to $0.86$ in the same setting. Even with DDPG, which generally lags behind SAC and TD3, PFR remains competitive; in the five-obstacle scenario with the Sneaker, PFR outperforms EPR ($0.81$ vs. $0.74$). These results highlight the benefit of preserving spatial structure through CNN-based encoding rather than 
flattening or relying on explicit pose lists. When comparing EPR and PFR more broadly, a clear trade-off emerges between efficiency and scalability. EPR converges faster, typically reaching stable performance after about $100$k training steps, but requires a fixed obstacle count and thus a separate policy for each scenario. In contrast, PFR requires longer training ($250$k steps) but produces a single policy that generalizes to arbitrary numbers and configurations of obstacles, making it more suitable for deployment in unstructured environments. This demonstrates that while EPR is efficient for controlled settings, PFR provides a scalable and robust representation for safe throwing in cluttered environments.

Although our policies achieve high success rates in simulation, we observed three recurring sources of failure:  

\noindent \textit{(i) Collisions with obstacles:} In cluttered scenarios, some throws result in trajectories that intersect with one or more obstacles. This typically occurs when the policy selects joint angles or release timings that generate shallow parabolic arcs. Upon contact, the object either bounces unpredictably or drops short of the basket.  

\noindent \textit{(ii) Underthrows and overthrows:} Another common failure arises when the throwing release timing is incorrectly predicted. In these cases, the object either falls short of the basket (underthrow) or overshoots it entirely (overthrow). While less frequent in obstacle-free conditions, these errors become more common as the environment grows more cluttered and valid throwing windows shrink.  

\noindent \textit{(iii) Suboptimal exploration in cluttered scenes:}  With three or more obstacles, the solution space narrows considerably, making policy optimization more challenging. Occasionally, the policy converges prematurely to a safe but suboptimal strategy, such as consistently throwing away from obstacles but also missing the basket.  

These failure modes highlight the inherent difficulty of learning robust throwing policies 
in highly constrained environments.
% , and motivate further work on exploration strategies and adaptive trajectory modulation to improve reliability in simulation.

\begin{figure}[!t]
    \centering
    \includegraphics[width=\linewidth]{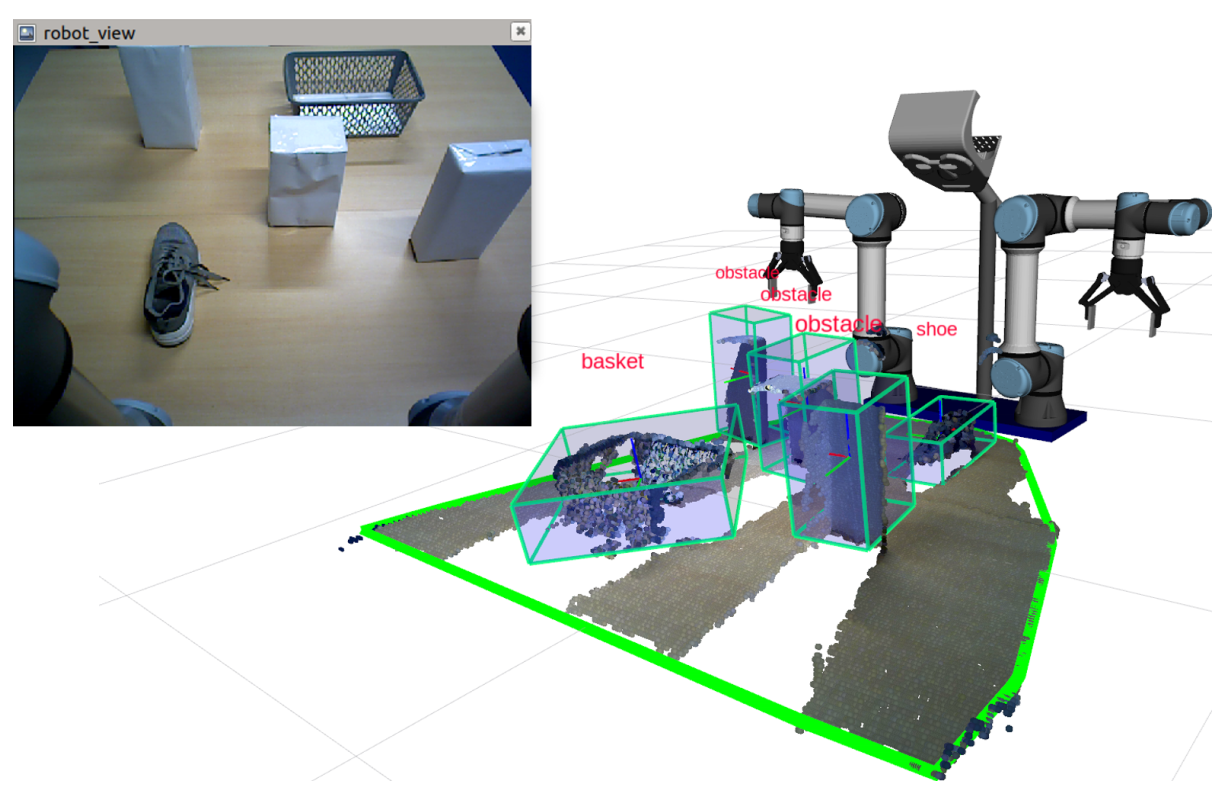}
    \caption{Example output of our real-robot perception system, showing object detection and recognition results.}
    \label{fig:real_perception}
    \vspace{-5mm}
\end{figure}

\subsection{Real Robot Experiments}
\begin{figure}[!t]
    \centering
    \includegraphics[width=\linewidth]{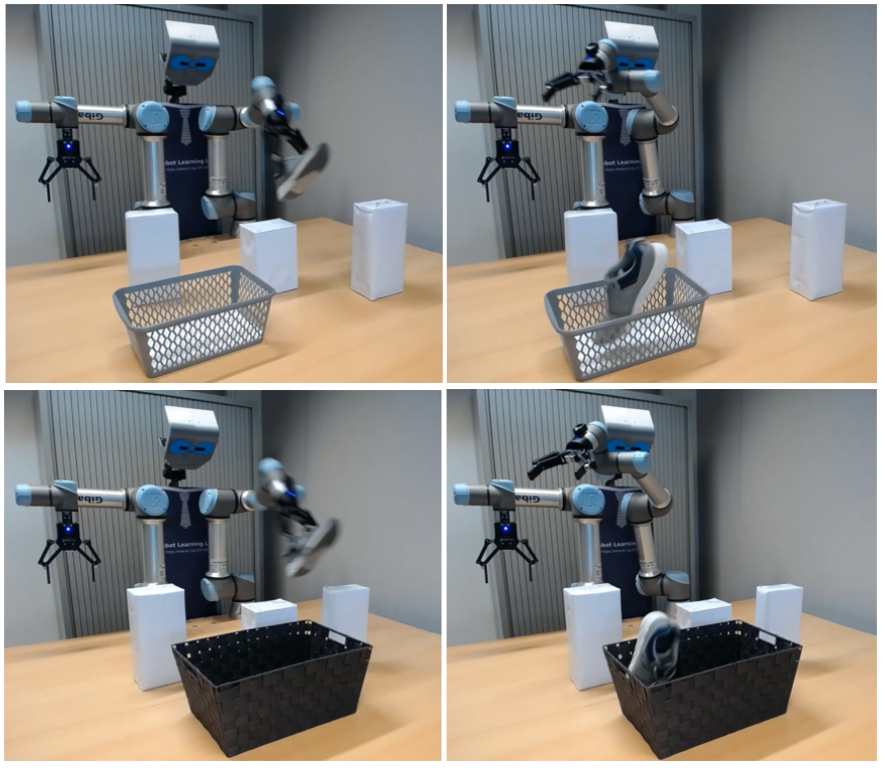}
    \caption{Throwing a sneaker object into the basket while avoiding obstacles: (\textit{top row}) an experiment with a basket size similar to the training phase; (\textit{bottom row}) an experiment with a basket size 1.5 times larger than the one used during training.}
    \vspace{-4mm}
    \label{fig:real_normal_and_large_basket}
\end{figure}

To directly compare the transferability of EPR and PFR, we evaluated both approaches on the real robot with three obstacles using the SAC algorithm. Each policy was trained in simulation and deployed zero-shot on the real robot without fine-tuning. For each configuration, we performed $20$ throwing trials using a sneaker as the throwable object and tested with both a basket of comparable size 
to the one used in simulation and a larger basket approximately $1.5\times$ wider (see Fig.~\ref{fig:real_normal_and_large_basket}). The model was provided with the same state information as in simulation; however, all perception data were obtained from an integrated perception pipeline 
described in~\cite{kasaei2018towards,kasaei2023mvgrasp,kasaei2019interactive}, 
which performs object detection, recognition, and pose estimation in real time. 
Figure~\ref{fig:real_perception} illustrates an example output of the perception system, showing accurate detection of multiple objects and their corresponding 6D poses.  
\begin{table}[!b]
\vspace{-3mm}
    \centering
    \caption{Real-robot throwing success rates (20 trials per setting) with SAC under EPR and PFR.}
    \resizebox{\linewidth}{!}{
    \begin{tabular}{|c|c|c|c|}
        \hline
        Basket Size & \#Trials & SAC-EPR & SAC-PFR \\
        \hline
        Normal (sim-sized) & 20 &  0.60 (12/20)&  0.70 (14/20)\\
        Large (1.5$\times$) & 20 &  0.70 (14/20) & 0.90 (18/20) \\
        \hline
    \end{tabular}}
    \label{tab:real_results}
\end{table}

Table~\ref{tab:real_results} summarizes the success rates. 
With the normal basket, PFR achieves higher accuracy ($70\%$) than EPR ($60\%$), 
demonstrating better robustness to perception noise and release-time variability. 
The advantage of PFR is even more pronounced with the larger basket, 
where it consistently reaches $90\%$, while EPR remains was $70\%$). 
These results confirm the scalability advantage of PFR: whereas EPR requires explicit encoding of each obstacle and thus is sensitive to pose estimation errors, PFR maintains a fixed state representation and generalizes more effectively to real-world clutter.  While success rates are slightly lower than in simulation, the relative trend between EPR and PFR remains consistent, 
highlighting the robustness of sim-to-real transfer.

By comparing the obtained results, we draw two key conclusions. 
First, the learned policies transfer effectively from simulation to the real robot, achieving high accuracy even with unseen throwable objects such as the sneaker. Second, performance is influenced by the target basket geometry: larger baskets mitigate the effects of perception noise and release-time variability, leading to higher success rates. Together, these findings confirm the feasibility of deploying the proposed method in real-world scenarios and demonstrate that our approach enables both safe and effective throwing in cluttered environments.

% \subsection{Failure Cases}
Despite the overall high success rates, we observed three recurring sources of failure during the real-robot experiments.  
% \cred{we can add a figure for failure cases}

\noindent \textit{(i) Inaccurate object pose estimation}: This refers to the challenge of precisely determining the position and orientation (pose) of the object in the robot's environment~\cite{deng2020self}.  This lag in tracking accuracy can lead to errors in predicting the object's trajectory during the throw, resulting in landing the object near, but not inside, the target basket.

\noindent  \textit{(ii) Delay in gripper command execution:} This issue arises due to the way the gripper is controlled. Unlike direct control, where the user can open and close the gripper instantly, it is governed by the robot's controller, adding a multi-step process. First, the user sends a command to the robot's controller, which then relays it to the gripper. However, the timing of this process is subject to network conditions and the robot's current operational state. Consequently, there can be a delay in the execution of gripping commands, potentially affecting the precision and timing of the object release during the throw.
\cite{liu2024tube} proposed an approach for addressing such releasing time uncertainty.

\noindent \textit{(iii) Unstable grasp pose selection:} A grasp that is stable for pick-and-place manipulation is not necessarily optimal for throwing, as the grasp configuration directly influences dynamics, object orientation, and resulting trajectory ~\cite{newbury2023deep,ten2017grasp}. Moreover, if the selected grasp pose is not stable or secure, it can lead to difficulties in maintaining a firm grip on the object during the throwing motion. These instabilities can result in deviations from the desired trajectory, contributing to failures in accurately delivering the object to the target basket.

These identified factors collectively contribute to the challenges faced in achieving successful object throws, in real robot experiments. Addressing these issues is crucial for improving the accuracy and reliability of the throwing process.

\section{Conclusion}
In this work, we addressed the problem of safe object throwing in cluttered environments, 
where a robot must deliver objects into a target basket while avoiding collisions with obstacles. 
We explored two complementary state representations: the Explicit Pose Representation (EPR), 
which directly encodes obstacle positions but requires a fixed obstacle count, 
and the Potential Field Representation (PFR), which provides a scalable, fixed-size encoding 
independent of the number of obstacles. To enhance the effectiveness of PFR, 
we introduced a CNN-based encoder that preserves the spatial structure of the potential field, 
enabling the policy to learn obstacle boundaries and basket attraction gradients more effectively.  

Through extensive simulation experiments, we showed that SAC consistently outperforms TD3 and DDPG, 
achieving high success rates across both EPR and PFR. Importantly, PFR not only closes the performance gap with EPR but, in many cases, surpasses it, while retaining the scalability advantage of handling arbitrary obstacle configurations. Real-robot experiments further confirmed the feasibility of our approach: policies trained in simulation transferred effectively to the physical robot, achieving up to $90\%$ success with unseen objects in cluttered scenarios.  

Overall, this work demonstrates that potential field representations combined with reinforcement learning offer a practical and robust solution for safe robotic throwing. By balancing accuracy, scalability, and transferability, our approach brings robotic throwing a step closer to real-world deployment in logistics, manufacturing, and service applications.
Looking ahead, several exciting directions remain open. One avenue is to extend the framework to dynamic multi-robot settings, where one robot throws an object, and another robot is responsible for catching it, or dual-arm coordination for throwing larger or heavier objects.
Another promising avenue is to extend our framework to dynamic environments with other policy learning approaches, such as Diffusion Policy or Diffusion Policy Optimization (DPPO), where both obstacles and targets may move unpredictably.

{
\small
\bibliographystyle{IEEEtran}
\bibliography{references}
}

% roslaunch ur_e_gazebo RL.launch gui:=false basic_controller:=1
% python3 rl_tossing_object_with_potential_field_cnn.py 
% gzclient

% tensorboard --logdir learnedPolicies/log_/

\end{document}